\documentclass{article}

\usepackage[preprint, nonatbib]{neurips_2024}

\usepackage[utf8]{inputenc} 
\usepackage[T1]{fontenc}    
\usepackage{hyperref}       
\usepackage{url}            
\usepackage{booktabs}       
\usepackage{amsfonts}       
\usepackage{nicefrac}       
\usepackage{microtype}      
\usepackage{xcolor}         

\usepackage{graphicx}
\usepackage{makecell}
\usepackage{bm}

\title{Interactive Simulations of Backdoors in Neural Networks}

\author{%
Peter Bajcsy\\
National Institute of Standards and Technology\\
 Gaithersburg, MD 20899\\
  \texttt{peter.bajcsy@nist.gov} \\
  \And
Maxime Bros\\
National Institute of Standards and Technology\\
 Gaithersburg, MD 20899\\
  \texttt{maxime.bros@nist.gov} \\
}

\begin{document}

\maketitle

\begin{abstract}
This work addresses the problem of planting and defending cryptographic-based backdoors in artificial intelligence (AI) models. 
The motivation comes from our lack of understanding and the implications of using cryptographic techniques for planting undetectable backdoors under theoretical assumptions in the large AI model systems deployed in practice. 
Our approach is based on designing a web-based simulation playground that enables planting, activating, and defending cryptographic backdoors in neural networks (NN). Simulations of planting and activating backdoors are enabled for two scenarios: in the extension of NN model architecture to support digital signature verification and in the modified architectural block for non-linear operators. 
Simulations of backdoor defense against backdoors are available based on proximity analysis and provide a playground for a game of planting and defending against backdoors. The simulations are available at \href{https://pages.nist.gov/nn-calculator}{https://pages.nist.gov/nn-calculator}
\end{abstract}

 \section{Introduction}
 
This work presents interactive simulations of planting, activating, and defending backdoors in fully-connected neural networks (NN). Backdoors are adversarial attacks on artificial intelligence (AI) systems that are planted by assuming control over training data,  test data, or source code \cite{Apostol2024}, \cite{langford2024architectural}, \cite{boberirizar2022architectural}. The attacker's goal with backdoor poisoning is to attack the integrity of AI models, which makes the AI model untrustworthy \cite{Apostol2024}.  The backdoor attack is successful when input into a poisoned AI model results in a wrong prediction with respect to the clean AI model prediction. 

Cryptographic methods can be used for integrity attestation of training data, test data, and AI models during different phases of an AI-model life cycle \cite{roy2023survey}, especially in life-critical applications such as medicine \cite{cryptomedicine2021}. The motivation for our work comes from the implications of using cryptographic techniques for planting and activating undetectable backdoors \cite{goldwasser2022planting}. There is a lack of understanding of how theoretical assumptions behind cryptographically undetectable backdoors \cite{goldwasser2022planting} \cite{dimitrov2022provably} scale to the AI models used in practice. Furthermore, there is a lack of understanding of how AI model robustness methods \cite{carlini2017evaluating}  \cite{madry2019deep} \cite{raghunathan2020certified} can defend AI model predictions in the presence of undetectable cryptographic backdoors.
Backdoor and data poisoning simulations aim to test backdoor propagation hypotheses, validate backdoor robustness methods, and educate scientists using AI models.
The ability to operate with neural networks and two-dimensional (2D) dot patterns is the same as with numbers in math calculators, which provides educational and research benefits to security scientists via explorations of NN and input data configurations. Simulations of backdoor and data poisoning are valuable for the AI security community preparing and solving Trojan detection challenges, for example, the Intelligence Advanced Research Projects Activity (IARPA)-sponsored TrojAI challenge \cite{IARPA2020} or the Trojan detection challenge of the NeurIPS conference \cite{trojandetection2022}.  

The problem of interest is a class of cryptographic architectural backdoors in AI non-linear activation functions.  The architectural backdoors have been introduced by Bober-Irizar et al. \cite{boberirizar2022architectural} and expanded by Langford et al. \cite{langford2024architectural}. 
 In a backdoored AI model, a malicious attacker can make imperceptible changes to the provided inputs or create a priori known input and trigger the backdoor behavior at the attacker's will. The inputs are formed by knowing a secret key that flips the sign of a numerical output from the backdoored first layer as outlined for undetectable backdoors by Goldwasser et al. \cite{goldwasser2022planting}. However, the output propagation from the backdoored first layer might not flip the AI model prediction with the increasing number of layers and with the final layer applying robustness methods \cite{madry2019deep} against adversarial attacks. 
This interplay between poisoned AI models via datasets (Trojaned AI models) and Trojan detectors was studied by  Sahabandu et al. \cite{sahabandu2024game}, Wu et al. \cite{Wu_2020}, and Carlini and Wagner \cite{carlini2023llm} but has not been fully explored for AI models with backdoors.     

The challenges of designing an interactive simulation of cryptographic backdoors lie in the many complexities of current AI models, which demand significant computational resources and prevent the simulations from being interactive.  
Furthermore, the challenges are in various cryptographic techniques and non-trivial solutions to creating inputs to trigger the cryptographic backdoors. 
Another challenge is the lack of adherence to a common terminology  \cite{Apostol2024} in scientific literature. For example, it is hard to determine based on the ``backdoor'' keyword of publications and GitHub repositories \cite{guo2019tabor}, \cite{aiguardian2023}, \cite{pan2023asset} whether methods are referring to data poisoning or source code poisoning according to the adversarial attack taxonomy \cite{Apostol2024}. 

\begin{figure}
\centerline{\includegraphics[width=12cm,keepaspectratio]{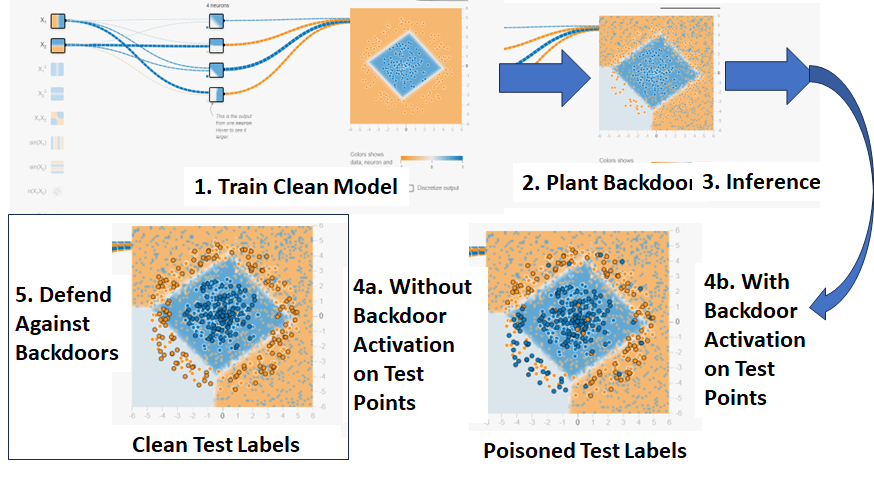}}
\caption{An overview of planting and activating a backdoor in a trained neural network. The simulation playground enables training a two-class fully-connected neural network (NN) with input features derived from 2D points, planting a checksum-based backdoor in the NN model, and activating the backdoor based on the knowledge of a secret key.}\vspace*{-5pt}
\label{objective}
\end{figure}

Our approach to this problem is based on designing a web-based simulation playground that enables planting, activating, and defending cryptographic backdoors with the overarching workflow illustrated in Figure~\ref{objective}.
 The web-based simulation playground is an extension of the neural network calculator \cite{bajcsy2021} and the neural network playground \cite{Smilkov2017}. The simulations are constrained to multiple 2D input data, ten features extracted from inputs, 64 fully-connected nodes (eight nodes per layer, eight layers), and two predicted classification labels visually labeled as blue and orange - see Figure~\ref{objective}. 
Among the slew of cryptographic methods for ensuring data integrity, we chose a simple checksum implementation out of many checksum methods \cite{GeeksforGeeks2023}.
This method follows black- and white-box AI models for planting undetectable backdoors \cite{goldwasser2022planting}. 
Finally, ``AI backdoors'' are implemented by controlling the source code \cite{Apostol2024}.
To minimize the source code modifications of an AI model,  our simulations use the NN model non-linear activation functions for planting a backdoor since the modifications do not change the model architecture (graph nodes and edges) and only insert a few lines of code with the checksum computation function call and one if statement. 
The feasibility of planting backdoors is explored in the Discussion section. 
   
Previous work on hiding AI model architectural backdoors is overviewed by Bober-Irizar et al. \cite{boberirizar2022architectural} and Langford et al. \cite{langford2024architectural}.
 The approaches include hiding backdoors in quantization or augmentation software routines \cite{ma2023quantization} or modifying the loss function of AI model optimization \cite{lossFunction2021}.  Furthermore, there have been reports of attacks on the PyTorch library in 2022 \cite{pytorch2022}, exploited vulnerabilities in the continuous integration and continuous development scripts for the PyTorch packages in 2024 \cite{pytorch2024}, and re-registered AI models in the Huggingface model repository in 2023 \cite{noy2023}.
 Our work is related to the previous work by planting backdoors via AI model modifications while focusing on the non-linear activation functions of AI nodes and cryptographic types of backdoors. Our work is also inspired by the backdoors that are provably undetectable \cite{goldwasser2022planting}, the adversarial examples that are robust to real-world perturbations \cite{dimitrov2022provably}, and the models that are resistant to adversarial attacks \cite{madry2019deep}.

In a black-box AI model scenario, we use the digital signature verification from \cite{goldwasser2022planting}  and the non-linear operators (a.k.a. activation functions) of each node in an AI model for planting architectural backdoors. In a white-box AI model scenario, we explore hypotheses about backdoor injections that could avoid a simple backdoor detection via source code inspection, for instance, by training the AI model with backdoors, saving the trained model weights, and re-loading the clean AI model with poisoned weights.  

Our contributions are in
\begin{enumerate}
\item designing an interactive web-based simulation framework for exploring checksum-based backdoors in fully-connected, small-scale, NN models with a variety of input features,
\item deriving input triggers that activate the backdoors planted in non-linear activation functions of the first layer of an NN model and 
\item enabling simulations of the game between adversaries planting backdoors and users defending against the backdoors in small-scale NN models.  
\end{enumerate}

 \section{Methods}
 
 In our work, an NN architectural backdoor is a hidden unauthorized functionality that can be activated by using a specific input (trigger) following the definitions of architectural software backdoors and data triggers  \cite{langford2024architectural} (Section 3.1). This section describes four functionalities in interactive simulations of planting checksum-based backdoors and defending against such backdoors. The first functionality is a simple checksum with a variable modulo value and target precision. The second functionality aims at activating backdoors in a digital signature verification example \cite{goldwasser2022planting}. The third functionality addresses how to plant and trigger (or activate) checksum-based backdoors in non-linear activation functions. The last functionality leverages the robustness methods against backdoor attacks based on proximity analyses. The methods accomplishing the four functionalities are presented next.

\subsection{Simple Checksum} 

While there are many different checksum functions \cite{GeeksforGeeks2023}, we use the simple checksum for simplicity during hypothesis testing.  Our specific definition of a simple checksum is provided in Equation~\ref{eq:csum}.

\begin{equation}
csum(v) = ( \sum_{i=1}^{L} s_{i} + (L_{MAX}-L) \times  48) \; mod \; m 
\label{eq:csum}
\end{equation}
where $csum(V)$ is the resulting checksum of a string representation $s$ of a value $v$, $s_{i}$ is the American Standard Code for Information Interchange (ASCII) value of the $i$-th character in $s$, 
$L = \min{(length(s), precision)}$ 
is the number of characters in $s$ limited by the precision, $L_{MAX}$ is the maximum number of digits per number, and $m$ is the modulo value.   
We extended the primary definition to include a precision, i.e., $L=\min{(length(s), precision)}$. Precision limits the number of most significant digits included in the sum to minimize the impact of numerical accuracy during double-precision floating point operations. Furthermore, we extended the basic definition in \cite{GeeksforGeeks2023} by padding the string representation with zeros (ASCII value is 48) to a fixed length so that all numbers contribute to the $csum$ value with the same number of digits. 

Inputs to simulations are double-precision floating point numbers in JavaScript that follow the international standard \cite{ieee754}. We convert a numerical input denoted as $v$ to its scientific notation and then to the string representation according to Equations~\ref{eq:rep}

\begin{equation}
v = coefficient 
\times
10^{exponent}; \; 
s = coefficient.toString() + exponent.toString()
\label{eq:rep}
\end{equation}
where $coefficient$ and $exponent$ include the minus sign as well.
Equation~\ref{eq:csum} is applied to the ASCII string representation of the coefficient and exponent values computed in Equation~\ref{eq:rep} with
 $L_{MAX}^{coefficient} = 24$, 
 $L_{MAX}^{exponent} = 4$, 
 $m=256$, and  
 $precision=15$. 
 The  $L_{MAX}$ value also includes a sign. Other configuration options are possible since modular arithmetic is commutative for addition $(a+b)\; mod \; m = (b+a) \; mod \; m$ and multiplication $(a \times b) \; mod \; m = (b \times a) \; mod \; m$.

\subsection{Backdoor in Checksum-Based Digital Signature Verification} 

\begin{figure}
\centerline{\includegraphics[width=12cm,keepaspectratio]{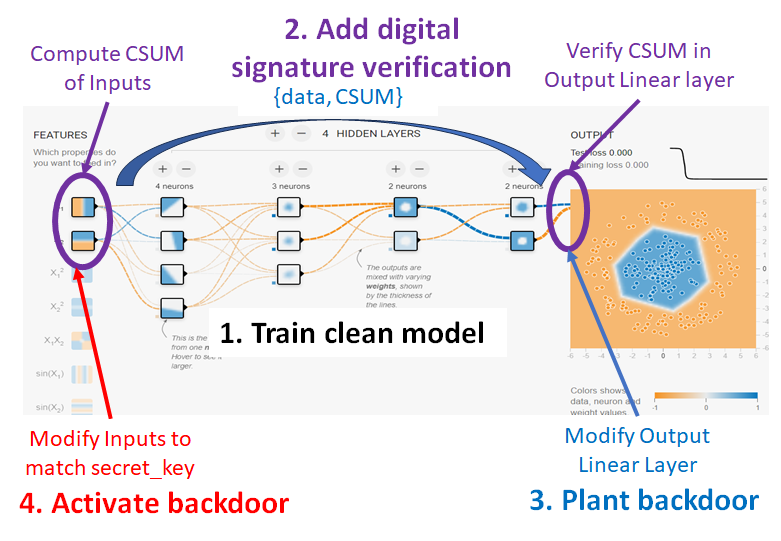}}
\caption{An overview of a neural network with digital signature verification described by Goldwasser et al.  \cite{goldwasser2022planting}. A backdoor is planted into the output linear layer to flip an output label for a checksum (CSUM) of the input value that matches a secret key.}\vspace*{-5pt}
\label{signature}
\end{figure}

The simulation of a backdoor attack on a checksum-based digital signature verification follows the description by Goldwasser et al.  \cite{goldwasser2022planting}. The implementation can plant a backdoor either in the output linear layer or the digital signature verification code. Our approach uses the output linear layer and is illustrated in Figure~\ref{signature}.   

\subsection{Checksum-Based Backdoor in Activation Functions of Model Nodes}
\label{sec:activate}


We use all NN nodes to plant a checksum-based backdoor in the rectified linear unit (ReLU) activation functions and backtrack the necessary minimal changes to the NN inputs of the first layer that flip the sign of the NN output from the first layer. The minimal changes to any input $x$ or $y$ coordinate are defined as ``visually'' imperceivable location changes or modifications to the least significant digits of the double-precision floating numbers representing $x$ and $y$ coordinates.  
 
The guaranteed activation solution is a pair of $x$ and $y$ inputs that satisfy  Equations~\ref{eq:sign}, \ref{eq:to}, \ref{eq:ti}, and \ref{eq:csum}, where $csum$ stands for a simple checksum operation and $sk$ stands for a secret key.
The coefficients $w_{1j}^{0}$ are the weights connecting $x$ with nodes indexed by $j$, the coefficients $w_{2j}^{0}$ are the weights connecting $y$ with nodes indexed by $j$, and the coefficients $b_{j}^{0}$ are the biases associated with each node $j$. The superscripts $0$ and $1$ refer to the layers.
 
  \begin{equation}\label{eq:sign}
    sign( \sum_{j=1}^{4}( w_{j1}^{1} \times TO_{j}^{1})  ) \neq 
    sign( \sum_{j=1}^{4} (w_{j1}^{1} \times (-TO_{j}^{1}))   ) 
\end{equation}

\begin{equation}\label{eq:to}
    TO_{j}^{1} = \max(0, TI_{j}^{1}) \; for \; j = 1,2,3,4 
\end{equation} 

\begin{equation}\label{eq:ti}
    TI_{j}^{1} = w_{1j}^{0} \times x + w_{2j}^{0} \times y + b_{j}^{0} \; for \; j = 1,2,3,4 
\end{equation} 

\begin{equation}\label{eq:csum}
    csum( TI_{j}^{1} ) = sk \; for \; j = 1,2,3,4 
\end{equation}

Unfortunately, a guaranteed activation solution based on a random search has a $m^{4}$ computational complexity for the single layer with four nodes. We confirmed this computational complexity by running the estimation of $x$ and $y$ one thousand times and counting the number of attempts until the solution was found. On average, our multiple estimation experiments found the $x$ and $y$ in $9617.562$ and $9742.891$  attempts for modulo $m=10$. 
In general, the random search has a $m^{n_{1}}$ computational complexity, where $n_{1}$ is the number of nodes in the first layer).

To achieve web-based interactivity, the modulo value $m$ must be adjusted based on the computer hardware.
Based on our benchmarks on a Dell laptop (Intel(R) Core(TM) i7-8850H CPU @ 2.60GHz), one evaluation of Equations~\ref{eq:csum} with a random pair $x$ and $y$ takes on average around $4.83 \times 10^{-5}$ 
seconds. Thus, the computational times for finding the activation values for $x$ and $y$ are: 
$m=10 \rightarrow 0.48\, s$,  
$m=20 \rightarrow 7.73\, s$, and
$m=30 \rightarrow 39.12 \, s$. 
These computational benchmarks suggest that interactive simulations cannot be delivered for modulo values $m \geq 20$.

\begin{figure}
\centerline{\includegraphics[width=12cm,keepaspectratio]{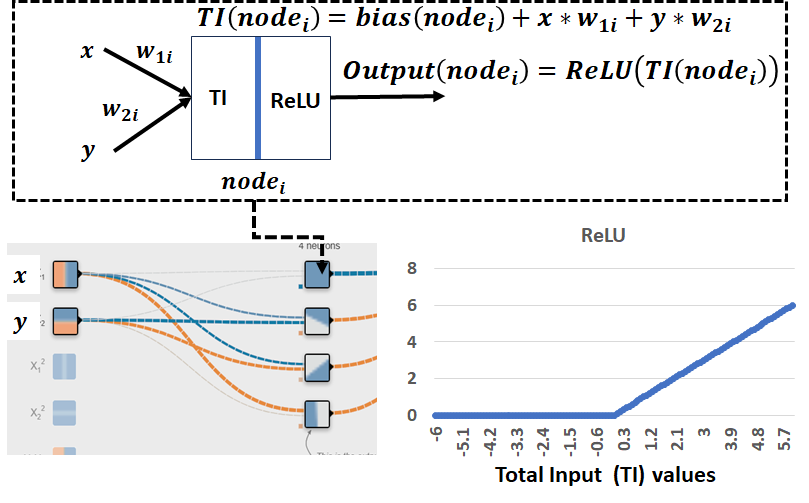}}
\caption{A schematic of computations in each NN node with rectified linear unit (ReLU). The total input of a node is computed for two input features, $x$ and $y$, as shown at the top. The NN first layer with four nodes, two inputs, and the ReLU activation function is shown at the bottom.}\vspace*{-5pt}
\label{node:relu}
\end{figure}

To address the interactivity requirement and use any modulo value, the backdoor activation in the simulation framework is constrained to one node in the first layer, which might not guarantee a flip of the output label but is highly likely to succeed.
The protocol for backtracking an input that activates the already planted backdoor is summarized as follows:
\begin{enumerate}

\item predict output label for a given $(x, y)$ input.

\item compute total input $TI_{i}$ to each node $i$ of the first layer and the sum of outgoing weights $SW_{i}$ of each node according to Equations~\ref{eq:ti} and \ref{eq:sw}.

\begin{equation}
TI_{i} = \sum_{j=1}^{N_{0}} w_{j,i} \times f_{j} + b_{i} 
\label{eq:ti}
\end{equation}
\begin{equation}
SW_{i} = \sum_{k=1}^{N_{2}} w_{i,k}
\label{eq:sw}
\end{equation}
where $f_{j}$ is one of the input features, 
$w_{j,i}$ is the weight between input feature $f_{j}$ and node $i$, 
$b_{i}$ is the bias associated with the node $i$,  
$N_{0}$ is the number of input features, and 
$N_{2}$ is the number of nodes in the second layer. 
Figure~\ref{node:relu} illustrates the computation of Total Input to a node with the ReLU activation function.

\item compute checksums of all $TI_{i}$ values and identify the optimal node $i_{SEL}$ to be activated which satisfies Equation~\ref{eq:f1} subject to the inequality in Equation~\ref{eq:f2}
 
\begin{equation}
i_{SEL} = arg\max_{i} SW_{i}
\label{eq:f1}
\end{equation}
\begin{equation}
s.t. \; |csum( TI_{i} ) - sk| < Th
\label{eq:f2}
\end{equation}
where $sk$ is the secret key value of the checksum, and $Th$ is the threshold for which one can modify digits of a total input value $TI_{i}$ to match the secret key $sk$. 

\item modify the digits of $TI_{i}$ iteratively from the least to the most significant digits within a precision-defined range so that Equation~\ref{eq:match} is satisfied.

\begin{equation}
|csum( \widehat{TI_{i_{SEL}}} ) - sk| = 0
\label{eq:match}
\end{equation}

\item reverse computation from $TI_{i_{SEL}}$ to the modified first input feature $\hat{f_{1}}$ according to Equation~\ref{eq:reverse}.

\begin{equation}
\hat{f_{1}} = \frac{\widehat{TI_{i_{SEL}}} - (\sum_{j=2}^{N_{0}} w_{j,i_{SEL}} \times f_{j} + b_{i_{SEL}} )}{ w_{j=1, i_{SEL}} }
\label{eq:reverse}
\end{equation}

\item inverse computation from the modified first input feature $\hat{f_{1}}$ 
to the $x$ or $y$ coordinate of the input point denoted as $\widehat{(x,y)}$. For example, one has to inverse any used input feature functions, such as
 $x^2$, $y^{2}$, 
 $x \times y$, $sin(x)$, $sin(y)$, $sin(x \times y)$,
 $sin(x^2 + y^2)$, or $0.5 \times (x+y)$.

\item verify the checksum value of $\widehat{TI_{i_{SEL}}}$ computed from the modified input coordinates 
$\widehat{(x,y)}$ being equal to a secret key. 

\item predict the NN output label using the modified input coordinates $\widehat{(x,y)}$ and report an attack success if the sign of a predicted label is opposite to the sign of the originally predicted label in Step 1.

\end{enumerate}

\begin{figure}
\centerline{\includegraphics[width=12cm,keepaspectratio]{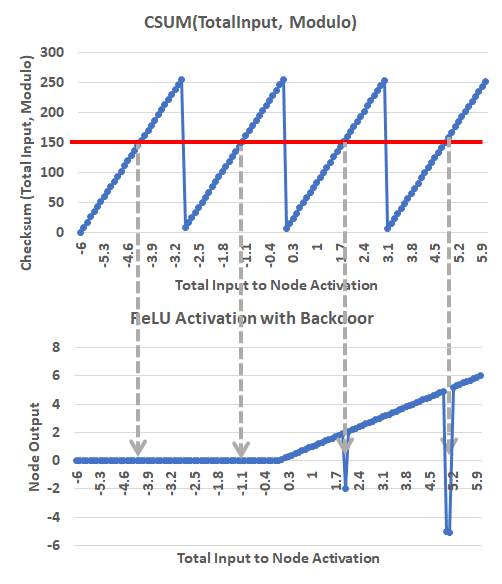}}
\caption{An illustration of the impact of a checksum function (top) with the secret key (red line) on the ReLU activation function (bottom). The arrows show the locations where the ReLU function will be affected by the checksum and the output sign will be flipped.}\vspace*{-5pt}
\label{relu:backdoor}
\end{figure}

This sequence is applied to a trained NN model and each input point. Depending on whether the conditions on input values of $(x, y)$ are satisfied, the input is modified to activate a backdoor in a selected node. $i_{SEL}$. 
An NN node with the backdoored ReLU function is illustrated in Figure~\ref{relu:backdoor}. An adversary can manipulate the modulo $m$, precision, and secret key values. Similar backdoors can be used with the Random Fourier Features \cite{Rahimi2007RandomFF} since the key idea is to introduce a periodic sign change of the output from a non-linear activation function.

\subsection{Defense Against Checksum-Based Backdoor}

The simulation follows the work by Madry et al. \cite{madry2019deep} that addresses a general question: ``How can we train deep neural networks that are robust to adversarial inputs?'' The defense approach \cite{madry2019deep} provides security guarantees against input perturbations bounded by $l_{\infty}$-bounded attacks. In general, AI defenses are always limited to a class of attacks (or input perturbations) and in a constant game with new attacks on data and AI models \cite{dimitrov2022provably}, \cite{goldwasser2022planting}.  
The simulation follows a protocol for defending against input perturbations as summarized next:
\begin{enumerate}
\item compute all pair-wise distances between blue points $d_{i,j}^{P}$, orange points $d_{i,j}^{N}$, and between blue and orange points $d_{i,j}^{NtoP}$. 

\item compute histograms of pair-wise distances: $\{d_{i,j}^{P}, \Delta R\} \longrightarrow \{ h_{k}^{P} \}$; 

\item select radius $R$ around an input point to be robust against input perturbations according to Equation~\ref{eq:robust}, where $k$ refers to the bin of a histogram of pair-wise distances.

\begin{equation}
R = \Delta R * (0.5 +  arg \max_{k}  \frac{h_{k}^{P} \times h_{k}^{N} }{h_{k}^{NtoP} } )
\label{eq:robust}
\end{equation}

\item count the number of blue and orange training points in the neighborhood of a test point defined by the radius $R$ and flip the test label if the count of the opposite label dominates the test point neighborhood. 

\end{enumerate}

 \section{Results} 
 
 This section presents the functionalities of checksum-based backdoors added to the neural network calculator \cite{bajcsy2021}. We outline simulations of planting backdoors in digital signature verification and ReLU activation functions, as well as defending against such backdoors using proximity analysis.

 \subsection{Backdoor in Checksum-Based Digital Signature Verification} 
 
\begin{figure}
\centerline{\includegraphics[width=12cm,keepaspectratio]{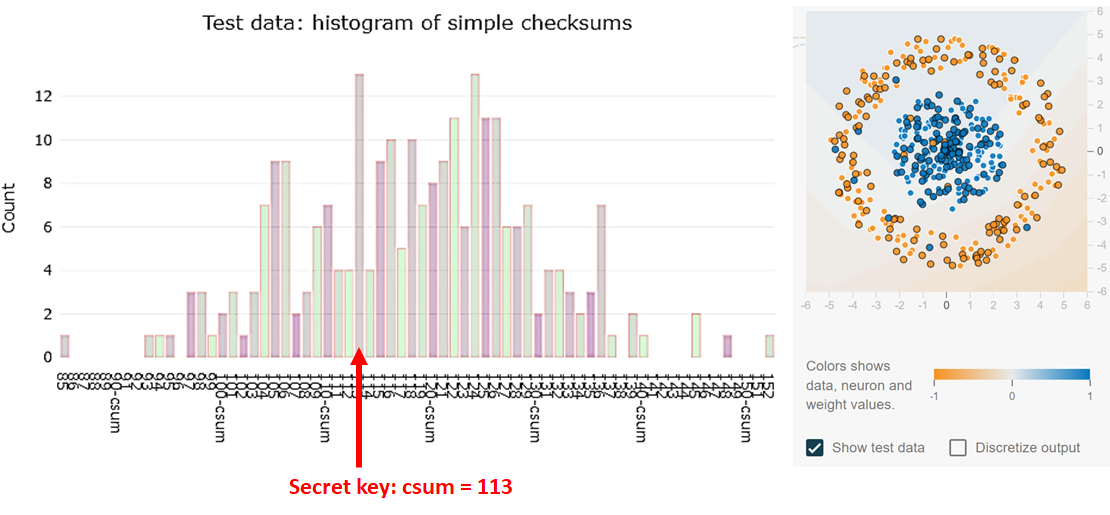}}
\caption{Histogram of checksum values for all test data points (left). Given a secret key (red arrow), all test points with the checksum equal to the secret key will flip their labels (13 labels are flipped for the secret key equal to 113).}\vspace*{-5pt}
\label{signature:hist}
\end{figure}

 By clicking on the ``CSUM Signature'' button, the simulation computes checksums of the x-coordinates of test data points and displays the histogram of those checksums. The test points with checksum values equal to the secret key flip their label and are shown by selecting the checkbox ``Show test data''.  
 Figure~\ref{signature:hist} shows the histogram and test data for the two classes of the ``doughnut'' input pattern.
 Repeated clicking on the ``CSUM Signature'' button will revert the flipped labels since the checksum values associated with each test data point do not change.

\subsection{Checksum-Based Backdoor in Activation Functions}

The attack using the activation functions can be executed in several steps. After training an NN network with $ReLU$ activation functions, the trained model is stored in memory using the NN calculator ``$NN\; M+$'' button. Planting the checksum-based backdoor is achieved by switching the ``Activation'' drop-down menu to $ReLU\_CSUM$ and reloading the trained model coefficients by using ``$NN\; MR$''. Now, the NN model has weights of the trained clean model and backdoors in each $ReLU$ activation function associated with each node.
 The backdoor in one of the nodes of the first layer is activated by backtracking input features for a subset of test points.

Table~\ref{table:01} shows numerical results of backtracking for an NN with four nodes in the first layer and the inputs composed of $x$ and $y$ coordinates. The backdoor secret key was $sk=150$. For any configuration of planting and activating backdoors, the corresponding values, as shown in Table~\ref{table:01},  are displayed on the simulation web page below the NN graph.

\begin{table}
 \caption{Example of intermediate variables to backtrack the input values that trigger a backdoor in the first layer and flip the output label. The original values are in the left column, and the modified values are in the right column. The changes in the total input $TI$ and in the backtracked $x$ coordinate are \underline{$\bm{underlined \; and\; bold}$}.}
 \label{table:01}
 \centering
\begin{tabular}{|c | c | }
\hline
\makecell{$TI=$\\$2.96880940\underline{\bm{35902}}424$} &  \makecell{$\widehat{TI}=$\\$2.96880940\underline{\bm{69999}}424$} \\
\hline
\makecell{$csum(TI)=127$} &  \makecell{$csum(\hat{TI})=150$}\\
\hline
\makecell{$x=$\\$-0.2875996\underline{\bm{7672464454}}$} & \makecell{$\hat{x}=$\\$-0.2875996\underline{\bm{939029182}}$}  \\
\hline
 \makecell{$y=$\\$-0.08406918324183005$} & \makecell{$\hat{y}=$\\ $-0.08406918324183005$} \\
\hline
\makecell{$output=$\\$0.9999999691015791$} &   \makecell{$\widehat{output}=$\\$-0.9999999999934367$}  \\
\hline
\makecell{$label=1$ (Blue)} &   \makecell{$\widehat{label}=-1$ (Orange)}          \\
\hline
\end{tabular}\vspace*{8pt}
\end{table}

%

\subsection{Defense Against Checksum-Based Backdoor}

\begin{figure}
\centerline{\includegraphics[width=12cm,keepaspectratio]{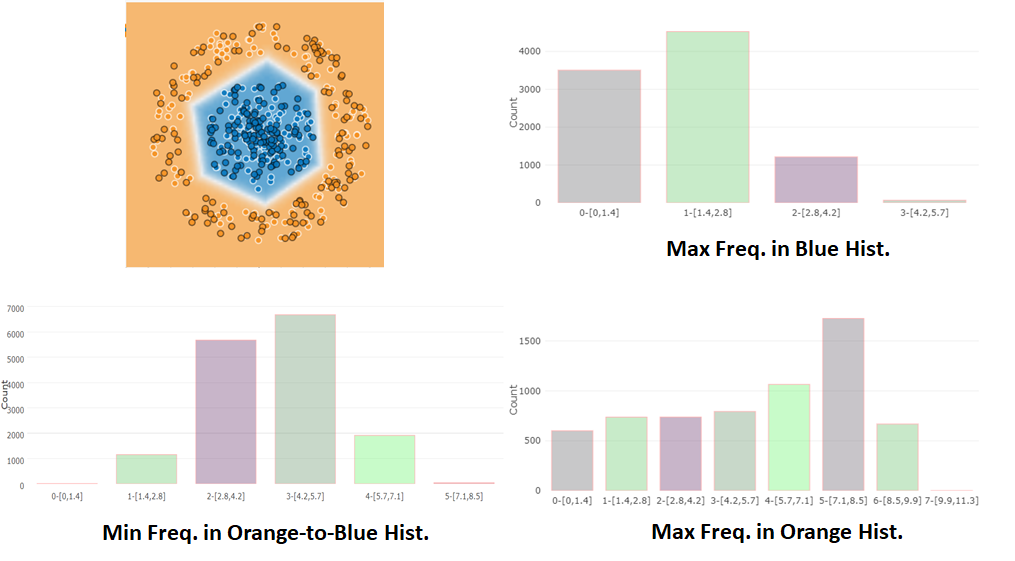}}
\caption{Histograms of the pair-wise distances of training data points between blue points (top right), orange points (bottom right), and between orange and blue points (bottom left) for the 2D points shown in the top left.
}\vspace*{-5pt}
\label{defense:hist}
\end{figure}

The simulation of defense against a checksum-based backdoor is executed in three steps. First, using the button ``CSUM Signature'', a test dataset will be contaminated with backdoored labels. Next, the pair-wise distances of training data points and their histograms can be computed using the button ``Label Proximity'' shown in Figure~\ref{defense:hist}. For the subset of training points, there are  6328 pairs of orange points, $9316$ pairs of blue points, and $15\,481$ pairs of orange to blue points, which are used to estimate a recommended radius to be the middle of the first histogram bin (selected $R=\frac{\sqrt{2}}{2}=0.707$). 
Finally, by clicking the ``Robust to Backdoor'' button, all test points are inspected based on their histogram of neighborhood labels, and a label is flipped if an opposite label dominates the neighborhood to the current one assigned to a test point.
 
 \section{Discussion}
 \label{sec:discuss}
 
 We discuss additional knowledge gained by simulating checksum-based backdoors in fully-connected neural networks. The discussion is divided into three major tasks of interest: planting, activating, and defending against backdoors.
 
 \subsection{Planting Backdoors}
From an attacking viewpoint, making a minimal change to the source code is critical. Contrary to a variety of attacks on the PyTorch library \cite{pytorch2022}, \cite{pytorch2024}, some scientists use PyTorch module modifications for solving emerging problems, for instance, the sustainable AI problem  \cite{wang2023bitnet}, \cite{ma2024era}. To address the sustainable AI problem, the nn.Linear layer is replaced with a BitLinear module in a BitNet architecture  \cite{wang2023bitnet}, which lowers the cost of training. While this demonstrates the feasibility of software modifications in the PyTorch library, the same approach can be misused by adversaries to plant a backdoor.
 
Another consideration is training an NN with a built-in checksum in the ReLU activation function. One can simulate such a case by switching the Activation menu to ``ReLU\_CSUM'' and observing the NN model convergence during training epochs. The convergence is not stable since the NN model cannot learn the checksum pattern. We did not modify the derivative part of ReLU since the checksum has no derivative needed during training.
It is possible to replace the checksum with a Random Fourier Feature (RFF) and its derivative as documented by Goldwasser et al. \cite{goldwasser2022planting}. While the implementation is for future work, one can currently simulate mathematically-related scenarios with input features being $sin(x)$, $sin(y)$, $sin(x \times y)$, or  $sin(x^2 + y^2)$.
 
One could envision two workflows for planting backdoors. The first workflow $WF_{1}$ consists of training a clean NN model, planting a backdoor in the NN model source code, and disseminating the NN model with a backdoor in the source code. The second workflow  $WF_{2}$ is composed of planting a backdoor in the NN model source code, training the NN model with the backdoor, and distributing a clean NN model with weights obtained by training the NN with the backdoor. The backdoor activation succeeds in  $WF_{1}$ but does not succeed in $WF_{2}$ because the total inputs to NN nodes have a random pattern of checksums over training epochs, and, hence, the NN model does not learn enough about checksums to be used as triggers. More analyses are needed to understand the training patterns of checksums fully.

 \subsection{Activation of Backdoors}

One possible adversarial approach is to collect checksums of total inputs $TI$ at a subset of NN nodes along the path to the output linear layer for selected test input (a set of features) and then form a vector or secret keys accordingly to control flips of output at any or all of the NN nodes. This is also a topic of future research and one part of simulations.
 
Additional discussion is relevant to the numerical precision, capacity of total inputs for modifications, and imperceptible modifications by humans and machines. As shown in Table~\ref{table:01}, the modification to $TI$ are smaller than $10^{-8}$ and larger than $10^{-14}$. These small changes can trigger a backdoor that is (to some degree) robust to machine precision and imperceptible to humans. However, quantization and half-precision hardware can minimize the capacity of total inputs for adversarial modifications (i.e., the number of available digits for modifications). In addition, for complex input features, for instance, $sin(x \times y)$ or  $sin(x^2 + y^2)$, retrieving $x$ or $y$ values using inverse functions might introduce numerical variability of the least significant digits.

\subsection{Defense Against Backdoors}
  
\begin{figure}
\centerline{\includegraphics[width=12cm,keepaspectratio]{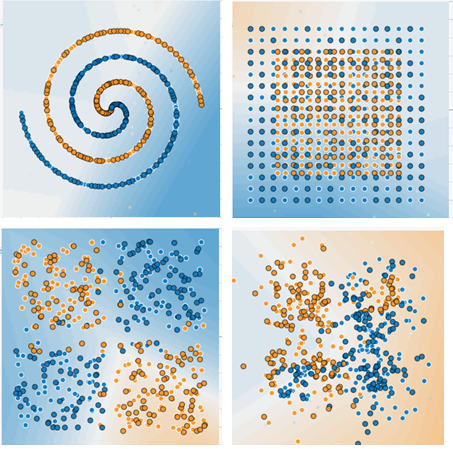}}
\caption{Examples of 2D dot patterns that violate the defense assumptions of input perturbations bounded by $l_{\infty}$-bounded attacks. Two classes are forming geometrical shapes of spiral and interleaved grid patterns (top) and being noisy and spatially overlapping (bottom). All patterns pose a challenge to defending against input perturbations. 
}\vspace*{-5pt}
\label{defense:patterns}
\end{figure}

The underlying assumption of the defense is that the training data classes are well-separated by a continuous boundary function in high-dimensional feature spaces.
Using mathematical terminology, 
the function representing the class boundaries must be a Lipschitz continuous function \cite{Lipschitz2007}, which limits sudden changes of labels in a feature space. 
We included in the simulation framework examples of 2D dot patterns that, together with noise level adjustments via the ``Noise'' sliding bar and a variety of data poisoning strategies invoked by the ``Trojan'' sliding bar, will violate the defense assumption. Figure~\ref{defense:patterns} illustrates four such examples of 2D dot patterns. For these patterns, local neighborhoods around labeled points must have a priori unknown shape and orientation (e.g., a spiral shape), satisfy a priori unknown property (e.g., constant pair-wise distances),  or be accompanied by a priori unknown class centroids and spatial densities (e.g., classes formed by multiple disconnected sets of points).

 \section{Conclusion}
 
 We have presented a design of an interactive online simulation framework for planting, activating, and defending checksum-based backdoors in fully-connected NN models. The simulations are available at \href{https://pages.nist.gov/nn-calculator}{https://pages.nist.gov/nn-calculator}
 
 The main benefits of the designed simulation framework lie in the interactivity and functionalities that become a playground for improving our understanding (and the implications of using) cryptographic techniques for planting and defending architectural backdoors in the large AI model systems deployed in practice. The simulations are built on the neural network calculator to operate with NN models and input datasets like with numbers in a math calculator. 
 
 While the simulations assume that adversaries have control over the source code of a model, the framework allows exploring scenarios with planting backdoors only during training, propagation of backdoored first layers to other layers, and defenses against a variety of complex 2D dot input patterns. These topics are subjects of future study.

\section*{Acknowledgement}
The funding for Peter Bajcsy was provided by the Intelligence Advanced Research Projects Activity (IARPA): IARPA-20001-D2020-2007180011. 

\section*{Disclaimer}
Commercial products are identified in this document to specify the experimental procedure adequately.
 Such identification is not intended to imply recommendation or endorsement by the National Institute of Standards and Technology, nor is it intended to imply that the products identified are necessarily the best available for the purpose.


\bibliographystyle{unsrt}
\bibliography{backdoor_arxiv}

\end{document}